%% file: main.tex
\documentclass{ieeeaccess}
\usepackage{cite}
\usepackage{amsmath,amssymb,amsfonts}
\usepackage{graphicx}
\usepackage{textcomp}

\usepackage{multirow, multicol}
\usepackage{kotex}
\usepackage{bm}
\usepackage{url}
\usepackage{amsmath}
\usepackage{xspace}
\usepackage{algorithm}
\usepackage{algpseudocode}
\usepackage{subfig}
\newcommand*{\Scale}[2][4]{\scalebox{#1}{$#2$}}
\makeatletter
\newcommand{\thickhline}{
    \noalign {\ifnum 0=`}\fi \hrule height 1pt
    \futurelet \reserved@a \@xhline
}
\makeatother

\usepackage{framed}
\usepackage{amssymb}
\usepackage{latexsym}
\algnewcommand\algorithmicswitch{\textbf{switch}}
\algnewcommand\algorithmiccase{\textbf{case}}
\algnewcommand\algorithmicassert{\texttt{assert}}
\algnewcommand\Assert[1]{\State \algorithmicassert(#1)}
\algdef{SE}[SWITCH]{Switch}{EndSwitch}[1]{\algorithmicswitch\ #1\ \algorithmicdo}{\algorithmicend\ \algorithmicswitch}
\algdef{SE}[CASE]{Case}{EndCase}[1]{\algorithmiccase\ #1}{\algorithmicend\ \algorithmiccase}
\algtext*{EndSwitch}
\algtext*{EndCase}

\def\BibTeX{{\rm B\kern-.05em{\sc i\kern-.025em b}\kern-.08em
    T\kern-.1667em\lower.7ex\hbox{E}\kern-.125emX}}
\begin{document}
\history{Date of publication xxxx 00, 0000, date of current version xxxx 00, 0000.}
\doi{10.1109/ACCESS.2017.DOI}

\title{s-DRN: Stabilized Developmental Resonance Network}
\author{\uppercase{In-Ug Yoon}\authorrefmark{1*}, 
\uppercase{Ue-Hwan Kim}\authorrefmark{1*}, \uppercase{Hyun Myung}\authorrefmark{1}, 
and \uppercase{Jong-Hwan Kim}\authorrefmark{1}
\IEEEmembership{Fellow, IEEE}.}
\address[1]{School of Electrical Engineering, KAIST (Korea Advanced Institute of Science and Technology), 291 Daehak-ro, Yuseong-gu, Daejeon 34141, Republic of Korea (e-mail: \{iuyoon, uhkim, johkim\}@rit.kaist.ac.kr, hmyung@kaist.ac.kr)}

\tfootnote{*The first two authors contributed equally to this work.}

\markboth
{Author \headeretal: Preparation of Papers for IEEE TRANSACTIONS and JOURNALS}
{Author \headeretal: Preparation of Papers for IEEE TRANSACTIONS and JOURNALS}

\corresp{Corresponding author: Jong-Hwan Kim (e-mail: johkim@rit.kaist.ac.kr).}

\begin{abstract}
Online incremental clustering of sequentially incoming data without prior knowledge suffers from changing cluster numbers and tends to fall into local extrema according to given data order. To overcome these limitations, we propose a stabilized developmental resonance network (s-DRN). First, we analyze the instability of the conventional choice function during node activation process and design a scalable activation function to make clustering performance stable over all input data scales. Next, we devise three criteria for the node grouping algorithm: distance, intersection over union (IoU) and size criteria. The proposed node grouping algorithm effectively excludes unnecessary clusters from incrementally created clusters, diminishes the performance dependency on vigilance parameters and makes the clustering process robust. To verify the performance of the proposed s-DRN model, comparative studies are conducted on six real-world datasets whose statistical characteristics are distinctive. The comparative studies demonstrate the proposed s-DRN outperforms baselines in terms of stability and accuracy.
\end{abstract}


\titlepgskip=-15pt

\maketitle

\input{1_introduction.tex}
\input{2_s_drn.tex}

\input{3_experiment.tex}
\input{4_conclusion.tex}

\section*{Acknowledgment}
This work was supported by Institute for Information \& communications Technology Promotion (IITP) grants funded by the Korea government (MSIT) (No.2016-0-00563, Research on Adaptive Machine Learning Technology Development for Intelligent Autonomous Digital Companion and No.2020-0-00440, Development of Artificial Intelligence Technology that Continuously Improves Itself as the Situation Changes in the Real World)

\bibliographystyle{IEEEtran}
\bibliography{reference}

\begin{IEEEbiography}[{\includegraphics[width=1in,height=1.25in,clip,keepaspectratio]{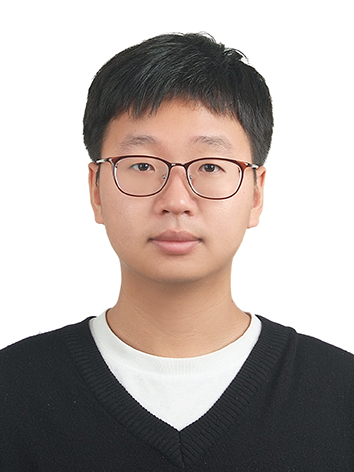}}]{In-Ug Yoon} received the M.S. and B.S. degrees in Electrical Engineering from Korea Advanced Institute of Science and Technology (KAIST), Daejeon, Korea, in 2018 and 2016, respectively. He is currently pursuing the Ph.D. degree at KAIST. His current research interests include anomaly detection, learning algorithms and computational memory systems.
\end{IEEEbiography}

\begin{IEEEbiography}[{\includegraphics[width=1in,height=1.25in,clip,keepaspectratio]{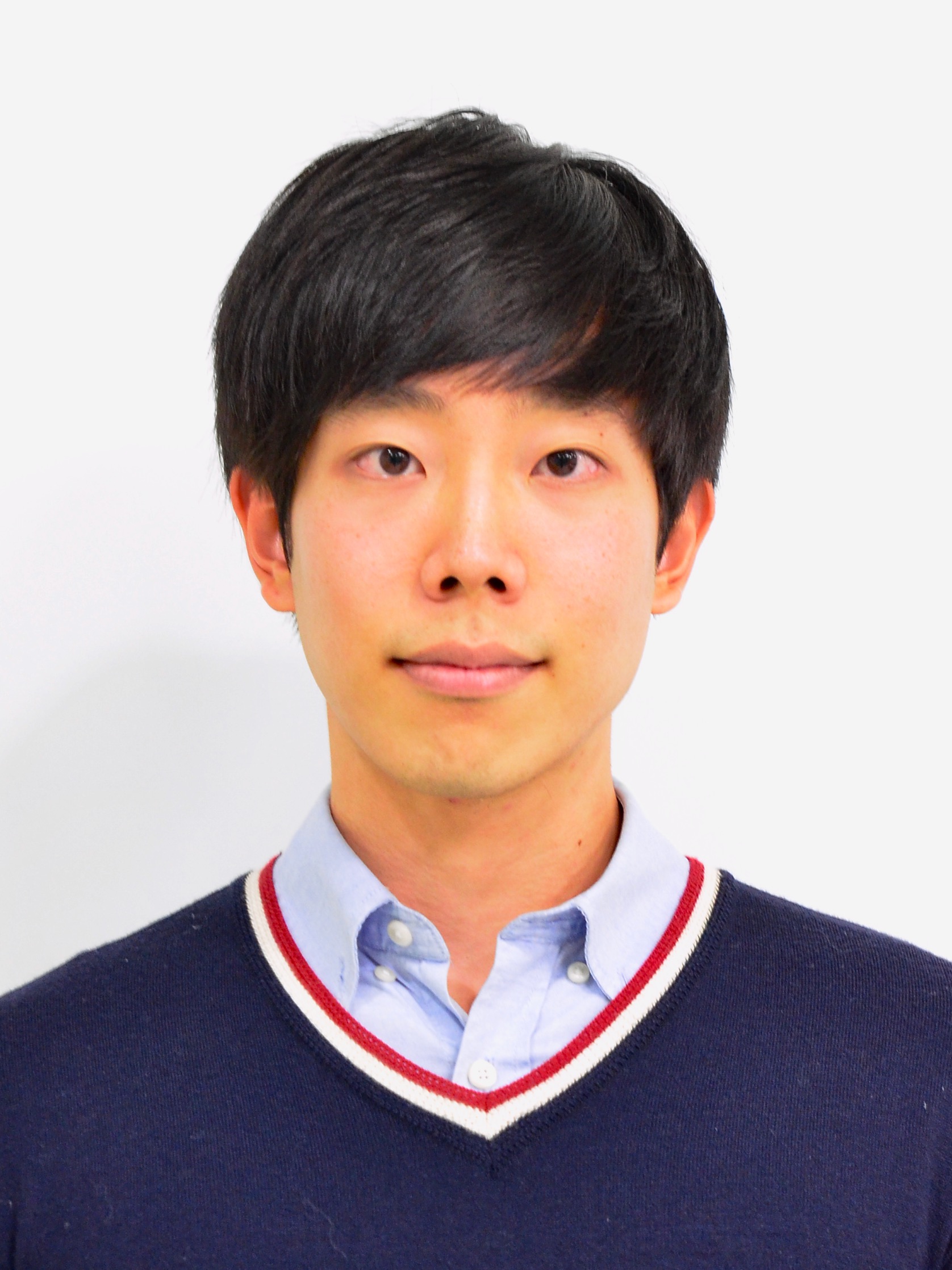}}]{Ue-Hwan Kim} received the Ph.D, M.S. and B.S. degrees in Electrical Engineering from Korea Advanced Institute of Science and Technology (KAIST), Daejeon, Korea, in 2020, 2015 and 2013, respectively. Since 2020, he has been with the School of Electrical Engineering, KAIST, Korea, as a postdoctoral researcher. His current research interests include visual perception, service robot, cognitive IoT, computational memory systems, and learning algorithms.
\end{IEEEbiography}

\begin{IEEEbiography}[{\includegraphics[width=1in,height=1.25in,clip,keepaspectratio]{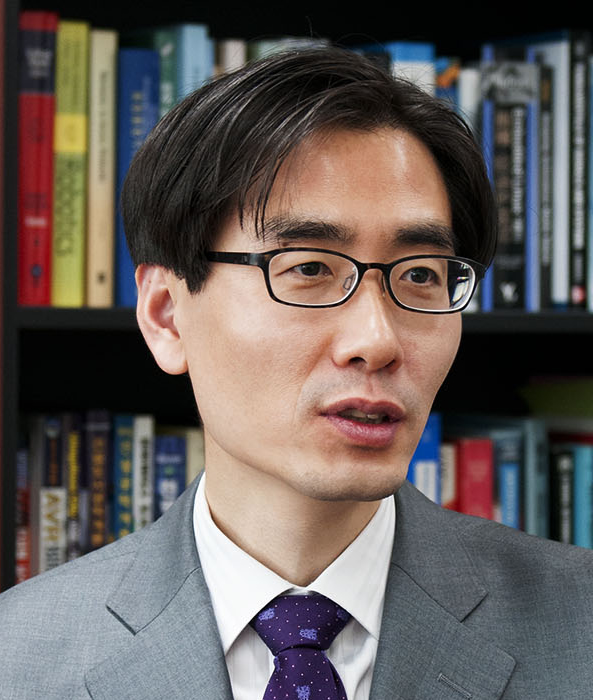}}]{Hyun Myung} (Senior Member, IEEE) received the B.S., M.S., and Ph.D. degrees from the Korea Advanced Institute of Science and Technology (KAIST), Daejeon, South Korea, in 1992, 1994, and 1998, respectively, all in electrical engineering. He was a Senior Researcher with the Electronics and Telecommunications Research Institute, Daejeon, from 1998 to 2002, CTO and Director of the Digital Contents Research Laboratory, Emersys Corporation, Daejeon, from 2002 to 2003, and a Principle Researcher with the Samsung Advanced Institute of Technology, Yongin, South Korea, from 2003 to 2008. From 2008 to 2018, he has been a Professor with the Department of Civil and Environmental Engineering, KAIST, where he is currently a Professor with the School of Electrical Engineering, KI-Robotics, KI-AI, and the Head of the KAIST Robotics Program. His current research interests include structural health monitoring using robotics, artificial intelligence, simultaneous localization and mapping, robot navigation, machine learning, deep learning, and swarm robots.
\end{IEEEbiography}

\begin{IEEEbiography}[{\includegraphics[width=1in,height=1.25in,clip,keepaspectratio]{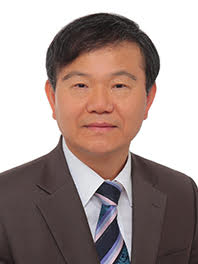}}]{Jong-Hwan Kim} (F'09) received the Ph.D. degree in electronics engineering from Seoul National University, Korea, in 1987. Since 1988, he has been with the School of Electrical Engineering, KAIST, Korea, where he is leading the Robot Intelligence Technology Laboratory as KT Endowed Chair Professor. Dr. Kim is the Director for both of KoYoung-KAIST AI Joint Research Center and Machine Intelligence and Robotics Multi-Sponsored Research and Education Platform. His research interests include intelligence technology, machine intelligence learning, and AI robots. He has authored 5 books and 5 edited books, 2 journal special issues and around 400 refereed papers in technical journals and conference proceedings.
\end{IEEEbiography}

\EOD

\end{document}

%% file: 1_introduction.tex
\section{Introduction}
\label{sec:introduction}
\PARstart{C}{lustering}, one of unsupervised learning algorithms, aims to group data instances into a number of categories. Clustering algorithms allow the analysis of data characteristics without prior knowledge, which can be applied to memory design \cite{wang2019k, nasir2017user,park2017deep,kim2018stabilized, ahmad2019survey}.
Clustering includes two main types of approaches: 1) batch learning and 2) online learning. The batch learning approaches, whose representative algorithms include $k$-means \cite{karlekar2019fuzzy} and GMM \cite{yegnanarayana2002aann}, are straightforward and simple to implement. However, they generally require a predefined cluster number from the user and all the training data to be given in advance. These features limit the application of batch learning algorithms in real-world applications where data are observed sequentially and continuously. 

On the other hand, online learning approaches can handle the varying number of clusters and incrementally process continuous data. Thus, in this paper, we focus on developing an effective online incremental clustering algorithm. Previous online learning approaches such as distance metric learning (DML) \cite{nguyen2019kernel} and self-organizing incremental neural network (SOINN) \cite{yu2019online} memorize all the given input and processing each input instance takes $O(n)$ computation. Fusion adaptive resonance theory (ART) \cite{masuyama2019topological, kim1996line, carpenter1998adaptive, grossberg2013adaptive} and Fuzzy ART \cite{akhbardeh2008towards, carpenter1991fuzzy, keskin2010fuzzy, da2019survey, da2020distributed} networks are efficient in the perspective of computation and memory usage, but they demand inputs to be normalized in the range of [0, 1] and the problem of node proliferation lingers \cite{jahromi2016sequential}. Developmental resonance network (DRN) \cite{park2019develop} has attempted to solve the two limitations, although its remedy for the normalization problem works for a certain range of input and it suffers from an inefficient grouping algorithm which is to solve the node proliferation problem.

To overcome the limitations mentioned above, we propose a stabilized developmental resonance network (s-DRN)\footnote{Source code available at \url{https://github.com/Uehwan/Incremental-Learning}}. The proposed s-DRN, free from the normalization problem, handles inputs with all scales and shows superior clustering performance. With the design of s-DRN, we solve the normalization problem by proposing a normalized node activation function. The node activation function proposed in DRN utilizes an exponential function and the activation value rapidly shrinks as the input scale increases. In such cases, DRN does not function as expected after a particular threshold scale. We propose to normalize input by the global weight vector which varies over time and the normalization problem disappears.

Next, we design a node grouping algorithm to alleviate the node proliferation problem. Since DRN and s-DRN allow unrestricted input scales, they cannot employ the complement coding scheme to prevent node proliferation and a node grouping algorithm for inhibiting node proliferation is essential. Three criteria, distance, intersection over union (IoU) and size criteria, are devised for the node grouping algorithm to effectively exclude unnecessary clusters from incrementally created clusters. In particular, we define and formulate the concept of IoU criterion for the node grouping algorithm. With the proposed IoU criterion, the node grouping algorithm becomes both scalable and stable in that the performance dependency on the vigilance parameter decreases. The proposed node grouping algorithm of s-DRN is computationally more efficient than that of DRN and s-DRN displays more effective clustering performance than conventional methods due to the proposed node grouping algorithm.

The remainder of this paper is structured as follows. Section II proposes the s-DRN model. Section III presents the experiment results with a thorough analysis. Concluding remarks follow in Section IV.

%% file: 2_s_drn.tex
\section{Stablized Developmental Resonance Network}
In this section, we delineate the computation flow of the s-DRN model. The whole process of s-DRN is summarized as Algorithm \ref{alg:s-DRN}.

\subsection{Global Weight Update}
s-DRN utilizes a global weight vector ($\bm{w}_{g} = [{}^{1}\bm{w}_{g}, ..., {}^{c}\bm{w}_{g}]$, where $c$ is the number of channels and ${}^{k}\bm{w}_g = [{}^{k}\bm{w}_{g1}; {}^{k}\bm{w}_{g2}]$) to cope with unknown scales of multi-channel inputs, which gets updated as follows:
\begin{equation}
\displaystyle
    {}^k\bm{w}_{g}^{\text{(new)}} = \begin{cases}
    \displaystyle
    {}^k\bm{x}_{i}, \:\:\:\:\:\:\:\, \text{if}\: i = 1 \\
    \displaystyle
    {}^k\bm{w}_{g}^{\text{(old)}}, \:\: \text{if}\: i \neq 1\: \text{and}\: d({}^k\bm{w}_{g}^{\text{(old)}}, {}^k\bm{x}_{i})=0\\
    \Scale[0.90]{
    (1-{}^kl_{g}){}^k\bm{w}_{g}^{\text{(old)}} + {}^kl_{g}([{}^k\bm{x}_{i} \land {}^k\bm{w}_{g1}^{\text{(old)}},{}^k\bm{x}_{i} \lor {}^k\bm{w}_{g2}^{\text{(old)}}]), 
    }\\
    \displaystyle
    \:\:\:\:\:\:\:\:\:\:\:\:\:\:\:\,\, \text{if}\: i \neq 1\: \text{and}\: d({}^k\bm{w}_{g}^{\text{(old)}}, {}^k\bm{x}_{i}) \neq 0
    \end{cases}
    \label{eq:DRN_global_weight_update}
\end{equation}
where ${}^k\bm{x}_i$ is the $i$-th step input of the $k$-th channel and ${}^kl _g \in (0,1]$ is the learning rate of ${}^k\bm{w}_g$. 

\subsection{Node Activation}
The input ${}^k\bm{x}_{i}$ activates the $j$-th node as follows:
\begin{equation}
\begin{split}
&T_j = \sum_{k=1}^{c}{}^k\gamma exp(-\frac{\alpha d({}^k\bm{x}_i,{}^k\bm{w}_j)}{{}^kM}),
\end{split}
\label{eq:sDRN_node_activation}
\end{equation}
where ${}^k\gamma$ is a contribution parameter and $\alpha$ is a slope parameter, $f(x) = exp(-\alpha x)$ is the choice function that normalizes the activation value $\bm{T}_{j}$ to $\bm{T}_{j} \in [0,1]$, and $d({}^k\bm{x}_{i},{}^k\bm{w}_{j})$ is the distance between ${}^k\bm{x}_{i}$ and the weight vector ${}^k\bm{w}_{j}$.

\begin{figure*}[t]
    \centering 
    \subfloat[Clustering result of DRN]{\includegraphics[width=7cm]{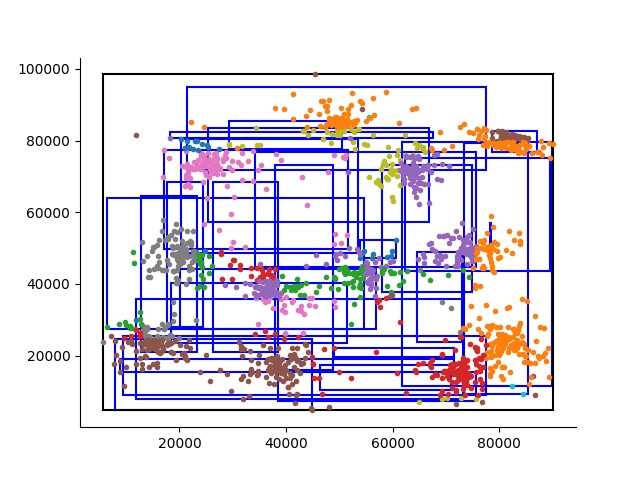}}
    \quad
    \subfloat[Clustering result of s-DRN]{\includegraphics[width=7.0cm]{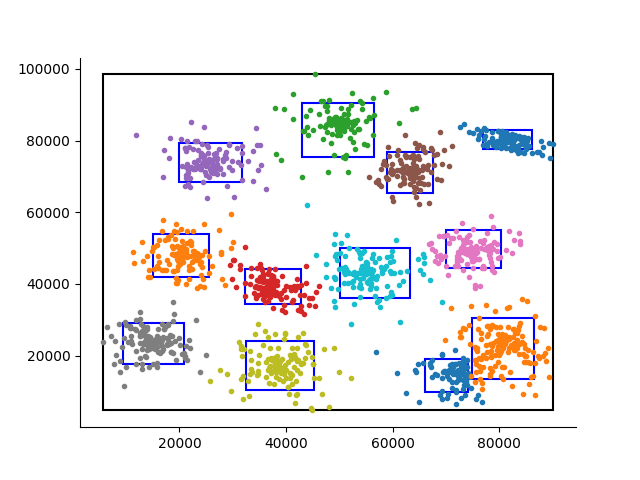}}
    \caption{Clustering results of DRN and s-DRN on example 2D synthetic data. Black lines indicate the global weight areas and blue lines represent each cluster weight boundary. As the input scale increases, DRN could not perform well due to the instability with node activation, while s-DRN shows robust performance with (\ref{eq:sDRN_node_activation}).}
    \label{fig:example_scalability}
\end{figure*}

For the exponential function to perform as a distance normalization function, ${}^k\gamma exp(-\alpha d({}^k\bm{x}_i,{}^k\bm{w}_j)) \geq \delta$ should be satisfied, where $\delta$ is the minimum value a processor supports. 
With the proposed activation function, s-DRN can handle all scales of input since ${}^k\gamma exp(-\frac{\alpha d({}^k\bm{x}_i,{}^k\bm{w}_j)}{{}^kM}) > \delta$ is invariably satisfied (Fig. \ref{fig:example_scalability}).

\subsection{Template Matching}
The template matching process identifies if the node with the largest activation value (say $J$-th node) resonates with the activity vector $\bm{x}_i$. First, the ratio $L_e({}^k\bm{x}_{i},{}^k\bm{w}_{J})=S_e({}^k\bm{x}_{i},{}^k\bm{w}_{J})/{}^kM_e$ between the two vectors ${}^k\bm{x}_{i}$ and ${}^k\bm{w}_{J}$ for each element $e$ ($e = 1,2, ..., {}^kz$ and ${}^kz$ is the dimension of the $k$-th channel) is calculated using the global diagonal vector ${}^kM= {}^k\bm{w}_{g2} - {}^k\bm{w}_{g1}$ of the $k$-th channel and the decision diagonal vector $S({}^k\bm{x}_{i}, {}^k\bm{w}_{J})= {}^k\bm{x}_{i} \vee {}^k\bm{w}_{J2} - {}^k\bm{x}_{i} \wedge {}^k\bm{w}_{J1}$ of the $J$-th node. Then, the resonance condition is defined as 
\begin{equation}
\begin{split}
&{}^km_J = \frac{{}^kz-|L({}^k\bm{x}_{i},{}^k\bm{w}_{J})|}{{}^kz} \geq {}^k\rho,
\end{split}
\label{eq:resonance_calculation}
\end{equation}
where ${}^km_J \in [0,1]$ is a resonance value, $L({}^k\bm{x}_{i},{}^k\bm{w}_{J})$ = $[L_e({}^k\bm{x}_{i},{}^k\bm{w}_{J})|e=1,...,{}^kz]$, and ${}^k\rho \in [0,1]$ is a vigilance parameter.

\subsection{Template Learning}
If the $J$-th node has resonated in the template matching process, the weight ${}^k\bm{w}_{J}$ gets updated by
\begin{equation}
\begin{split}
&{}^k\bm{w}_{J}^{\text{(new)}} = (1-{}^kl){}^k\bm{w}_{J}^{\text{(old)}} + {}^kl ([{}^k\bm{x}_{i} \wedge {}^k\bm{w}_{J1}^{\text{(old)}}, {}^k\bm{x}_i \vee {}^k\bm{w}_{J2}^{\text{(old)}}]),
\end{split}
\label{eq:template_learning}
\end{equation}
where ${}^kl \in (0,1]$ is the learning rate of the $k$-th channel. 

\subsection{Node Grouping}

The proposed node grouping algorithm mitigates the performance instability attributed to data input order and the dependency on vigilance parameters. The proposed node grouping process compares the activated cluster with nearby clusters when an input vector arrives and groups a pair if two clusters in the pair satisfy three criteria: distance, IoU and size criteria. The three conditions are examined over all channels and all the channels should satisfy each condition for the examination of the next condition.

\begin{figure}
	\centering
	\includegraphics[width=0.4\textwidth]{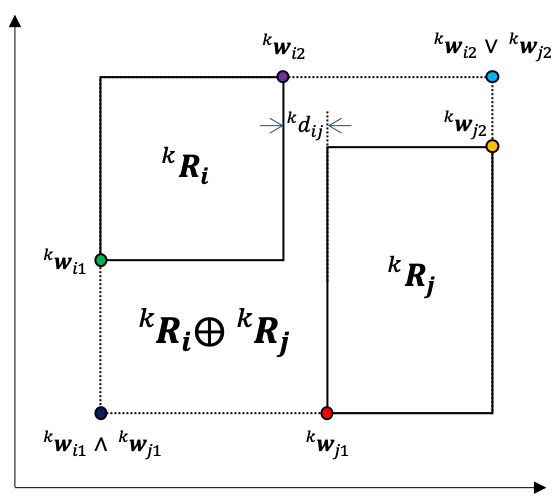}
	\caption{Visualization of the proposed node grouping algorithm.}
	\label{fig:grouping_example}
\end{figure}

For the formulation of the criteria, let $R_i$ and $R_j$ denote a pair of neighboring clusters (Fig. \ref{fig:grouping_example}). The weight vectors representing each cluster for the $k$-th channel are:
\begin{equation}
\begin{split}
&{}^kR_i = \{{{}^k\bm{w}}_i = ({{}^k\bm{w}}_{i1}; {{}^k\bm{w}}_{i2}) | k=1,...,c\} \quad \text{and} \\
&{}^kR_j = \{{{}^k\bm{w}}_j = ({{}^k\bm{w}}_{j1}; {{}^k\bm{w}}_{j2}) | k=1,...,c\},
\end{split}
\end{equation}
where $c$ is the number of channels and semicolon represents concatenation. We define the distance vector between a pair of clusters as
\begin{equation}
\begin{split}
{\bm{d}}_{ij} = ({}^1d_{ij}, {}^2d_{ij}, ... , {}^cd_{ij}),
\end{split}
\end{equation}
where each element of the vector is defined as
\begin{equation}
\begin{split}
{}^kd_{ij} = \min(|{{}^k\bm{w}}_i - {{}^k\bm{w}}_j| \wedge |({{}^k\bm{w}}_{i2};{{}^k\bm{w}}_{i1})- {{}^k\bm{w}}_j|),
\end{split}
\end{equation}
where $min()$ operator chooses the minimum element of a vector. The proposed distance criterion is
\begin{equation}
\begin{split}
\max({\bm{d}}_{ij}) < 1-\rho,
\end{split}
\end{equation}
where $max()$ operator chooses the maximum element of a vector and $\rho$ is a vigilance parameter for the template matching. Note that in s-DRN, $\rho$ is used instead of ${}^k\rho$ due to unnecessity of vigilance parameter for each channel.

We propose the IoU criterion since the distance criterion can become loose and combine all the clusters when a low valued vigilance parameter is used. The IoU criterion tests if the hypothetically grouped cluster could encompass the two compared clusters with the least extension. This guarantees the grouped cluster does not occupy un-investigated feature space substantially. The below represents the hypothetically grouped cluster for the $k$-th channel:

\begin{equation}
\begin{split}
{}^kR_i \oplus {}^kR_j &=\\
\{ {}^k{\bm{w}}_{i \oplus j} &= ({}^k{\bm{w}}_{i1} \wedge {}^k{\bm{w}}_{j1};({}^k{\bm{w}}_{i2} \vee {}^k{\bm{w}}_{j2})) | k = 1,...,c\}.
\end{split}
\end{equation}
For each category cluster, we define the volume of the $k$-th channel as
\begin{equation}
\begin{split}
{}^kV_j = \prod_{y=i}^{dim({}^k{\bm{w}}_{j})}({}^k{\bm{w}}_{j2}-{}^k{\bm{w}}_{j1})_y.
\end{split}
\end{equation}
Next, we define the IoU criterion for the $k$-th channel as

\begin{equation}
\begin{split}
IoU({}^kR_i, {}^kR_j) = \frac{{}^kV_i+{}^kV_j}{{}^kV_{i \oplus j}} > \tau,
\end{split}
\label{eq:sdrn_iou}
\end{equation}
where $\tau$ determines the final threshold for the grouping process. The range of IoU value is in [0,2] and we set $\tau$ as 0.85 through empirical study. 

The size criterion limits the maximum size of a category cluster. Excessively large clusters resulted from node grouping hinder the normal template matching process. Thus, we limit the size of a cluster. The maximum size of the $j$-th cluster for the $k$-th channel ($|{}^k\bm{w}_{j2} - {}^k\bm{w}_{j1}|$) is limited to $^{k}M(1-\rho)$, which is congruent to (\ref{eq:resonance_calculation}).

\subsection{Computational Efficiency Analysis}
The computational complexity of fusion ART on which DRN is based is $T(n) = hzn$, where $h$ is the number of categories, $z$ is the dimension of the input, and $n$ is the number of data samples. With its grouping algorithm, the computational complexity of DRN becomes
\begin{equation}
\begin{split}
T(n) = (hz + v(v-1)/2)n + mq,
\end{split}
\end{equation}
where $m$ and $q$ are the average numbers of global weight updates and connected category pairs, respectively.

On the other hand, the computational complexity of s-DRN is

\begin{equation}
\begin{split}
T(n) = (hz)n + hz.
\end{split}
\end{equation}
The increase of computation $hz \approx O(h)$ with s-DRN is minuscule compared to that of DRN which is $(1/2)v(v-1)n + mq \approx O(n)$.


\begin{algorithm}
\caption{s-DRN algorithm}
\textbf{Input}: Training data $x_i$; hyper-parameters: $\rho,\tau,l$\\
\textbf{Output}: Clustered data
\begin{algorithmic}[1]
\State // Handling the global weight vector
\State Calculate the distance $d(x_i, w_g)$
\If{$d(x_i, w_g) \neq 0$}
    \State Update the global weight vector $w_g$ by (\ref{eq:DRN_global_weight_update})
\EndIf
\State
\State // Checking resonance
\State Calculate the activation value $T_j(x_i)$ using (\ref{eq:sDRN_node_activation})
\State Select the node $J$ with the largest activation value
\State Calculate the resonance value $m_J$ using (\ref{eq:resonance_calculation})

\If{$m_J \geq \rho$}
\State Update the weight $w_J$ by (\ref{eq:template_learning})
\State break
\Else
\State Generate a new cluster, and initialize the value of the weight to $x_i$
\EndIf
\State
\State // Performing the node grouping algorithm
\For{{$j =1$ to $h$}}
        \State Calculate criteria values between clusters $J$ and $j$ 
        \State Achieve grouping flag $flag_{gr}$ from grouping criteria
\EndFor
\If{$flag_{gr}$}
    \State Group cluster J and the indexed cluster
\EndIf

\end{algorithmic}
\label{alg:s-DRN}
\end{algorithm}

%% file: 3_experiment.tex
\section{Experiments}
In this section, we illustrate the experiment setting for performance verification and establish the effectiveness of the proposed s-DRN model.

\subsection{Experiment Setting}

\subsubsection{Datasets}
We retrieved six real-world benchmark datasets from the UCI machine learning repository\footnote{\url{https://archive.ics.uci.edu/ml/index.php}}: Balance scale, liver disorder, blood transfusion service center, bank note authentication, car evaluation, and wholesale customers datasets. We attentively selected the set of datasets so that each dataset fairly differs in the size of the dataset, the number of clusters, input dimensions, and scale ranges.

\subsubsection{Metrics}
For quantitative analysis, we employed three performance metrics. First, Davies-Bouldin index (DBI) \cite{coelho2012automatic} estimates the ratio of within-cluster scatter to between-cluster separation as follows:
\begin{equation}
\begin{split}
&DBI = \frac{1}{K}\sum_{k=1}^{K}{\max_{j \neq k}{(\frac{\sigma_k+\sigma_j}{d(\mu_k,\mu_j)})}},
\end{split}
\end{equation}
where $K$ is the cluster number, $\mu_y$ is the center point of cluster $y$, $\sigma_x$ is the average distance of every element $x$ in a cluster to $\mu_x$ and $d(\mu_k,\mu_j)$ is the distance between $\mu_k$ and $\mu_j$. The lower value of DBI indicates higher clustering performance.

Next, clustering purity (CP) \cite{mukherjee2019clustergan} matches each output cluster to the ground-truth cluster as follows:

\begin{equation}
\begin{split}
&CP(\Omega, C) = \frac{1}{N}\sum_{k=1}^{K}{\max_{j}{|w_k \cap c_j|}},
\end{split}
\end{equation}
where $\Omega = \{w_1,w_2,...,w_K\}$ is the set of clusters and $C = \{c_1,c_2,...,c_J\}$ is the set of ground-truth classes. Since a large number of clusters can bias CP, we complemented CP with normalized mutual information (NMI) \cite{knops2006normalized} which is defined as

\begin{equation}
\begin{split}
&NMI(\Omega,C) = \frac{2 \times I(\Omega;C)}{[H(\Omega)+H(C)]},
\end{split}
\end{equation}
where $H$ is entropy and $I(\Omega;C)$ is mutual information between $\Omega$ and $C$. Both CP and NMI lie in the range of [0, 1], where a larger value implies higher performance.

\subsubsection{Baselines}
For comparative studies, we employed three baseline algorithms: $k$-means, GMM and DRN \cite{park2019develop}. $k$-means and GMM are two representative batch-based clustering algorithms and the number of clusters should be given in advance. On the other hand, DRN and s-DRN are online learning algorithms and the number of clusters increases in an incremental manner.

\subsubsection{Implementation Detail}




To reduce the effect of randomness, we conducted each experiment 100 times and report the average and the standard deviation of each metric. In addition, each experiment received the data instances in different orders. For $k$-means and GMM, we split the datasets into train and test sets with the ratio of 5:5. We set the ratio, which showed the best performance for $k$-means and GMM after sweeping the ratio from 1:9 to 9:1. Moreover, we provided $k$-means and GMM with the ground-truth cluster numbers.

For DRN and s-DRN, we sequentially input data instances and did not provide the ground-truth cluster numbers. We set one vigilance parameter, $\rho$ for both DRN and s-DRN. The optimal parameters were obtained using the follow metric:
\begin{equation}
\begin{split}
&T_j = \sum_{k=1}^{c}(\xi_{d}(DBI) + \xi_{c}(1-CP) + \xi_{n}(1-NMI))
\end{split}
\label{eq:clustering_global_metric}
\end{equation}
where $\xi_{d}$, $\xi_{c}$ and $\xi_{n}$ are reciprocals of standard deviations of DBI, CP and NMI, respectively. We swept the vigilance from 0.1 to 0.9 and found the best value (0.7 and 0.5 for DRN and s-DRN, respectively) according to (\ref{eq:clustering_global_metric}). We use one vigilance parameter since vigilance parameter cannot be fine-tuned in the real-world setting. In the real-world setting, no prior knowledge of dataset is generally given and data instances come sequentially.


\subsection{Results and Analysis}

\begin{table*}[!t]
\renewcommand{\arraystretch}{1.1}
\caption{Results of Comparative Studies}
\centering
\begin{tabular}{l | c c c | c c c | c c c}
\thickhline
\multirow{2}{*}{\textbf{Algorithm}} & \multicolumn{3}{|c|}{\textbf{Balance Scale}} & \multicolumn{3}{|c|}{\textbf{Liver Disorder}} & \multicolumn{3}{|c}{\textbf{Blood Transfusion}}\\ 
\cline{2-10}
& \textbf{DBI} & \textbf{CP} & \textbf{NMI} & \textbf{DBI} & \textbf{CP} & \textbf{NMI} & \textbf{DBI} & \textbf{CP} & \textbf{NMI}\\
\hline
\multirow{2}{*}{\textbf{$k$-means}} & 1.6059 & 0.6805 & 0.1592 & 1.1229 & 0.3526 & \bf{0.1483} & \bf{0.4588} & 0.7620 & \bf{0.0535}\\
& (0.0266) & (0.0198) & (0.0510) & (0.1690) & (0.0176) & \bf{(0.0121)} & \bf{(0.0416)} & (0.0146) & \bf{(0.0048)}\\
\hline
\multirow{2}{*}{\textbf{GMM}} &  1.6562 & 0.6899 & 0.1383 & 1.7940 & 0.3364 & 0.1069 & 0.8218 & 0.7541 & 0.0130\\
& (0.0475) & (0.0265) & (0.0265) & (0.3177) & (0.0025) & (0.0180) & (0.0065) & (0.0004) & (0.0011)\\
\hline
\multirow{2}{*}{\textbf{DRN}} & 1.3065 & 0.6734 & 0.1459 & 1.0624 & 0.3426 & 0.0594 & 0.5851 & 0.7645 & 0.0264 \\
& (0.1546) & (0.0425) & (0.0280) & (0.5307) & (0.0037) & (0.0190) & (0.1452) & (0.0029) & (0.0081)\\
\hline
\multirow{2}{*}{\textbf{s-DRN}} & \bf{1.0707} & \bf{0.8137} & \bf{0.2572} & \bf{0.6951} & \bf{0.3642} & 0.1309 & 0.4802 & \bf{0.7679} & 0.0306 \\
& \bf{(0.0622)} & \bf{(0.0230)} & \bf{(0.0185)} & \bf{(0.1266)} & \bf{(0.0134)} & (0.0341) & (0.0609) & \bf{(0.0010)} & (0.0029)\\
\hline
\hline
\multirow{2}{*}{\textbf{Algorithm}} & \multicolumn{3}{|c|}{\textbf{Banknote}}& \multicolumn{3}{|c|}{\textbf{Car Evaluation}} & \multicolumn{3}{|c}{\textbf{Wholesale Customers}}\\
\cline{2-10}
& \textbf{DBI} & \textbf{CP} & \textbf{NMI} & \textbf{DBI} & \textbf{CP} & \textbf{NMI} & \textbf{DBI} & \textbf{CP} & \textbf{NMI}\\
\hline
\multirow{2}{*}{\textbf{$k$-means}} & 0.9871 &  0.7653 & 0.1976 & 1.8841 & 0.6944 & 0.1742 & 0.7585 & \bf{0.5818} & 0.1447\\
& (0.0570) & (0.0369) & (0.0560) & (0.0797) & (0.0120) & (0.0368) & (0.1000) & \bf{(0.0295)} & (0.0141)\\
\hline
\multirow{2}{*}{\textbf{GMM}} & 1.5829 & \bf{0.7754} & \bf{0.2861} & 1.9428 & 0.6963 & 0.1492  & 1.7571 & 0.5674 & \bf{0.2075}\\
& (0.3689) & \bf{(0.1465)} & \bf{(0.2059)} & (0.2084) & (0.0032) & (0.0374) & (0.3343) & (0.0174) & \bf{(0.0182)}\\
\hline
\multirow{2}{*}{\textbf{DRN}}& 0.9437 & 0.6040 & 0.0525 & 1.6907 & 0.7145 & 0.1404 & 2.3920 & 0.5065 & 0.0698\\
& (0.2176) & (0.0341) & (0.0459) & (0.2976) & (0.0176) & (0.0397) & (0.8174) & (0.0357) & (0.0556)\\
\hline
\multirow{2}{*}{\textbf{s-DRN}}& \bf{0.8734} & 0.7088  & 0.1630& \bf{1.1625} & \bf{0.8041} & \bf{0.2246} & \bf{0.3720} & 0.4975 & 0.0748\\
& \bf{(0.2124)} & (0.0633) & (0.0648) & \bf{(0.0571)} & \bf{(0.0179)} & \bf{(0.0148)} & \bf{(0.1930)} & (0.0170) & (0.0246)\\
\thickhline
\end{tabular}
\label{table:summary}
\end{table*}

Table \ref{table:summary} summarizes the results of comparative studies. s-DRN consistently displays superior performance over all six datasets achieving small values for DBI and large values for CP and NMI. We note that s-DRN outperforms $k$-means and GMM on average although $k$-means and GMM were given the ground-truth cluster numbers and the half of each dataset was given as a training set. The comparative studies corroborate that s-DRN guarantees satisfactory clustering performance in an online incremental manner compared to batch-based clustering algorithms. Moreover, the performance of s-DRN surpasses that of DRN over all six datasets, which verifies the effectiveness of the proposed node grouping algorithm.

\begin{figure}
	\centering
	\includegraphics[width=0.45\textwidth]{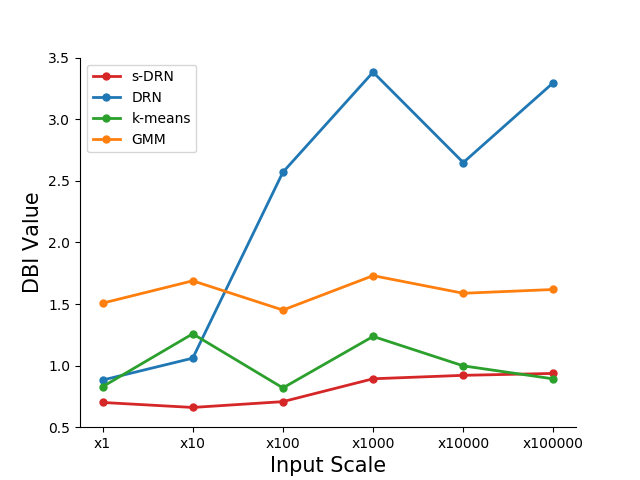}
	\caption{Effect of input scales on clustering performance. Elements of liver disorder dataset was scaled from $\times1$ to $\times100,000$ for clustering performance measurement.
    }
	\label{fig:real_scale_effect}
\end{figure}

Particularly, the performance gap between DRN and s-DRN is the largest for the wholesale customer dataset. The large input scale of the dataset interrupts DRN's activation function and its performance deteriorates sharply. The result of the wholesale customer dataset confirms that the proposed activation function truly resolves the normalization problem. Fig. \ref{fig:real_scale_effect} further investigates the effect of input scale on the clustering performance. We tested each algorithm on the liver disorder dataset and varied the scale from $\times1$ to $\times100,000$. The effect of input scale on other algorithms including s-DRN is insignificant while the performance of DRN gets sensitively affected.

\begin{figure*}
\centering
\subfloat[Result of DRN]{%
\includegraphics[width=0.45\textwidth]{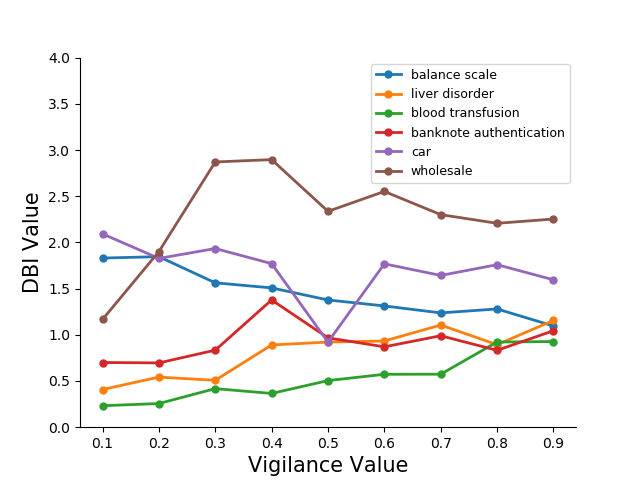}
}
\subfloat[Result of s-DRN]{%
\includegraphics[width=0.45\textwidth]{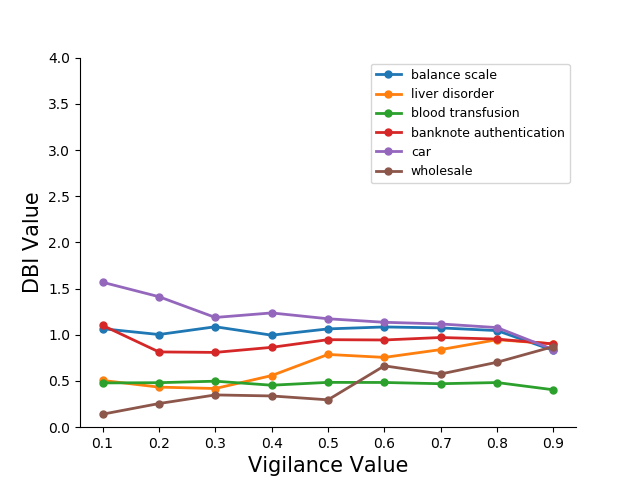}
}
\hfill
\caption{Effect of vigilance parameter on clustering performance for DRN and s-DRN. Vigilance parameter value was varied from 0.1 to 0.9 for performance measurement.}
\label{fig:vig_effect}
\end{figure*}

Fig. \ref{fig:vig_effect} illustrates the effect of the vigilance parameter on clustering performance for DRN and s-DRN. For all six datasets, we varied the vigilance parameter from 0.1 to 0.9 and observed the performance variation in DBI. As the figure exhibits, the clustering performance of s-DRN is stable over all vigilance values in all six datasets. However, the clustering performance of DRN strongly depends on the value of the vigilance parameter. The effect of the vigilance parameter for DRN and s-DRN establishes that the node grouping algorithm of s-DRN is more effective and efficient than that of DRN. For quantitative analysis, we report the averages of standard deviations of DBI scores for DRN and s-DRN, which are 0.307 and 0.143, respectively.

%% file: 4_conclusion.tex
\section{Conclusion}
In this paper, we proposed a resonance-based online incremental clustering network, s-DRN, which is a stabilized model of DRN. The proposed s-DRN model resolves the normalization problem remaining in conventional methods with the proposed activation function. Thus, s-DRN can effectively handle all input scales. Moreover, s-DRN equipped with the proposed node grouping algorithm becomes robust to variation of vigilance parameter, and the need for fine-tuning vigilance parameter disappears. In addition, the clustering performance improves with the proposed node grouping algorithm. A thorough examination of s-DRN through experiments on six real-world benchmark datasets established the effectiveness of s-DRN. We expect s-DRN can be applied to various real-world settings where no prior knowledge on sequentially incoming data is given.